\documentclass[runningheads]{llncs}
\usepackage[T1]{fontenc}
\usepackage{booktabs}
\usepackage{caption}
\usepackage{subcaption}
\usepackage{cite}
\usepackage[font={small}]{caption}
\usepackage[utf8]{inputenc}
\usepackage{hyperref}

\usepackage{graphicx}
\begin{document}
\title{MACHINE LEARNING FOR CLASSIFICATION OF ANTITHETICAL EMOTIONAL STATES}
\titlerunning{ML FOR CLASSIFICATION OF ANTITHETICAL EMOTIONAL STATES}
%
\author{Jeevanshi Sharma\inst{1} \and
Rajat Maheshwari\inst{2} \and
Yusuf Uzzaman Khan\inst{3}}
\authorrunning{J. Sharma et al.}
%
\institute{Department of Electrical Engineering, Aligarh Muslim University, Aligarh, India 202001\\ \email{sharma.jivanshi@gmail.com} \and
Department of Computing Science, Lovely Professional University, Punjab, India 144411\\
\email{rajatmaheshwari72@rediffmail.com}\\
\and
Department of Electrical Engineering, Aligarh Muslim University, Aligarh, India 202001\\
\email{yusufkhan.ee@amu.ac.in}}
\maketitle              
\begin{abstract}
Emotion Classification through EEG signals has achieved many advancements. However, the problems like lack of data and learning the important features and patterns have always been areas with scope for improvement both computationally and in prediction accuracy. This works analyses the baseline machine learning classifiers performance on DEAP Dataset along with a tabular learning approach that provided state-of-the-art comparable results leveraging the performance boost due to its deep learning architecture without deploying heavy neural networks. 

\end{abstract}
\section{Introduction}
The ability of a machine to understand its surrounding has become indispensable way to interact with technology. As it has touched all the areas of our life; it will be meaningful to create human-aware computing which can under-
stand the need of the user or can adapt to the surrounding situation. Emotion is a psycho-physiological process that is generated by the conscious or unconscious perception of an object or stimuli. Temperament, personality and disposition, and motivation are all factors to consider\cite{r1} when dealing with emotions. Emotions are equally important in everyday conversation. A simple affirmation can communicate sorrow, grief, anger, happiness, or even contempt depending on context; however, the complete meaning can only be comprehended via facial expressions, hand gestures, and other non–verbal communication tools. Because not all emotions can be visually seen and observed, emotion categorization in its entirety necessitates an examination of several characteristics and variables, including commonly sensed emotions such as happiness or rage in differing degrees and emotions that are difficult to comprehend by just observing. As a result, the relevance of emotion categorization is much more when it comes to mental health disorders and illnesses. Mental illnesses are critical to public health concerns because it contribute significantly to the worldwide epidemics and have a significant influence on people's social and economic well-being. Because of these raised concerns around the periphery of mental illnesses and the increase in their spread throughout populations, governments are now emphasizing the mental well-being of citizens for sustainability, economic growth, and productivity \cite{r2}. Timely detection of such mental illnesses and conditions is crucial for long-term benefits and well-being. Detecting emotional characteristics and intensities are second nature to humans and detecting them with facial recognition software has proven challenging and costly. And thus, most Human Computer Interface (HCI) systems lack emotional intelligence and are unable to interpret a detected emotion. Thus, an effective computing system is expected to fill this gap by detecting emotional cues through HCI systems, as well as synthesizing emotional responses \cite{r1}.
\\
Electroencephalography (EEG) is a technique for recording an electrogram of electrical activity on the scalp, which has been proven to represent the macroscopic activity of the brain's surface layer underneath. EEG is a clinical term that refers to the recording of the brain's spontaneous electrical activity across time using several electrodes implanted on the scalp.\cite{r5} Researchers in past have taken advantage of EEG signals to diagnose diseases, such as sleep disorders, epilepsy,\cite{r7} schizophrenia \cite{r8} etc. There exist two popular approaches to categorizing emotions - Discrete categorization \& Dimension based categorization. While discrete emotions are classified into six different categories - sadness, joy, surprise, anger, fear, and disgust, dimensional approaches classify emotions as one of the three - valence, arousal, and dominance. This dimensional approach is based on an observation that affective states are not independent, and are related to each other in a systematic manner. In this way, the dimensional based approach is able to represent a larger set of emotions. Dominance represents the scale of control ranging from without control to empowered as compared to valence (level of positivity or negativity) and arousal (level of excitement or apathy). Thus, several studies including this focus only on valence and arousal for simplicity and ease. Therefore, the problem of emotion recognition boils down to the permutation of classifying High Valence vs Low Valence and High Arousal vs Low Arousal.

EEG is considered reliable for emotion recognition as it detects the underlying brain activity. Use of EEG as an input modality has a number of advantages that make it suitable for use in real-life tasks including its non-invasive nature and relative tolerance to movement. The main focus of emotion recognition has been on feature extraction and classification. the critical step during detecting emotion from EEG signals is extracting features, which traditionally has been extracting temporal or spectral features from a fixed group of same EEG channels for all subjects.\cite{r9}  In studies, features like Discrete wavelet transform (DWT), Power Density Spectrum (PSD) ratio, Hjorth parameter, Energy and correlation features \cite{r11} have also been used to classify EEG emotion signals, into negative and positive emotion using SVM and KNN \cite{r12} classifiers.  In \cite{r10}, the authors proposed to eliminate noisy and redundant channel using PCA and ReliefF algorithms and then trained a Support Vector Machine using DEAP data-set which resulted into an accuracy of 81.7\%. In \cite{r13}, authors worked on DEAP data-set for verifying the activated regions of brain while experiencing a negative emotion. For a subset of the participant readings in the data-set, correlation for emotion corresponding to high arousal and low valence was observed in the right hemisphere of the brain.  In a comparative study \cite{r23} on various features of EEG for emotion recognition, PSD performed best as compared to other features like spectral power asymmetry, higher order spectra, higher order crossing, common spatial patterns and asymmetric spatial patterns for classifying two emotions - happy and unhappy. Other features extracted from combination of electrode are utilized too, such as coherence and asymmetry of electrodes in different brain regions \cite{r16} \cite{r17}and graph-theoretic features\cite{r18}.
\\
However, it is difficult in machine learning approaches to cover all the implied features simply by extracting them manually, given the complex formulations of time-domain and frequency-domain feature extraction methods. Also, EEG signals are susceptible to noise interference and it becomes difficult to extract the key time points or segments with high emotional correlation in the EEG sequence. To overcome it, researchers have used various end-to-end deep learning based classifiers. In \cite{r19} used convolution neural networks (CNN) for EEG decoding and visualization which have shown great potential when applied to end-to-end EEG based emotion recognition. In \cite{r20}, authors proposed a deep learning network based on Stacked Auto-encoder (SAE) to model the energy spectral density features of EEG signals. They also applied a PCA-based principal component covariant adaptive transformation algorithm for classifying emotion in arousal and valence dimensions. However, a typical end-to-end CNN fails due to the low generalization ability of the CNN on EEG data present in small sample sizes\cite{r21}. Limited available EEG data is one of the major root causes of challenges associated with feature extractions and model generalizability. The publicly available data-sets like SEED and DEAP have much smaller scale compared to data-sets available for Image or Text. Generating artificial data by applying a transformation from the original data is one of the conventional solutions to solving the data scarcity problem. various data augmentation methods have been applied to generate EEG data \cite{r22} \cite{r23} \cite{r24}, along with deep generative model based approaches \cite{r25}. Deep learning models can automatically end-to-end learn the abstract features from large scale raw samples, avoiding the engineering of feature extraction and feature selection. However, in the field of EEG signal recognition and classification, large-scale labeled EEG data-sets are limited and many EEG-based BCI (Brain Computer Interface) applications often require high real-time performance. 

In this study, the objective is to analyse the performance of different machine learning classifiers and a tabular learning approach on DEAP Data-set using Welch’s method (a part of Fourier analysis) to extract theta, alpha, beta and gamma spectral power for each electrode (32 channels). Secondly, accuracy scores across all the channels and different EEG regions have been evaluated against two levels of arousal and two levels of valence. The focus have been on producing a better accuracy scores without deploying deep neural networks.

\section{Dataset and Material}

\subsection{DEAP Dataset}
The DEAP database consists of the electroencephalogram (EEG) and peripheral physiological signals of 32 participants (16 men and 16 women, aged between 19 and 37, average: 26.9) recorded as each watched 40 one-minute long excerpts of music videos. The EEG signals, sampled at 512 Hz, were recorded from the following 32 positions (according to the international 10-20 positioning system, see Figure 1): Fp1, AF3, F3,
F7, FC5, FC1, C3, T7, CP5, CP1, P3, P7, PO3, O1,
Oz, Pz, Fp2, AF4, Fz, F4, F8, FC6, FC2, Cz, C4, T8,
CP6, CP2, P4, P8, PO4, and O2. The proposed music videos were demonstrated to induce emotions to different users \cite{r26} represented in the valence-arousal scale. Participants rated each video in terms of the levels of arousal, valence, like/dislike, dominance, and familiarity. The same videos had an online evaluation that could be used for comparison. For 22 of the 32 participants, frontal face video was also recorded. A novel method for stimuli selection is proposed using retrieval by affective tags from the last.fm website, video highlight detection, and an online assessment tool. An extensive analysis of the participants' ratings during the experiment is presented. Correlates between the EEG signal frequencies and the participants' ratings are investigated. Data were provided both as they were acquired (raw data) and in the pre-processed form. DEAP data-set is provided in .mat files with labels in the same format. To process it with our pipeline we have converted it .csv files for each subject. The provided data is down sampled at 128Hz but were collected at 512Hz.

\begin{table}
\centering
\caption{Description of DEAP Dataset}
\resizebox{\textwidth}{!}{%
\begin{tabular}{@{}lll@{}}
\toprule
\textbf{Index} & \textbf{Data}                          & \textbf{Value}                              \\ \midrule
1.             & Number of Participants                 & 32(16 Males and 16 Females)                 \\
2.             & Participant's Age                      & Between 19 and 37 (Avg. age 26.9 years)     \\
3.             & Stimuli Used                           & Visual Stimuli in the form of Music Videos  \\
4.             & Number of Videos                       & 40                                          \\
5.             & Video Duration                         & 60 seconds (1 minute each)                  \\
6.             & Recorded EEG Signals                   & Using 32-channel 512 Hz EEG                 \\
7.             & Other Signals Recorded                 & Peripheral Physiological Signals (GSR,      \\
8.             & Number of data points for each video   & 8064 values per electrode (8064 x 32)       \\
9.             & Total values recorded for each patient & 322560x32                                   \\
10.            & Total data points in complete DEAP     & (322650x32) x 32                            \\
11.            & Tag or labeling recording              & Recorded by every participant after         \\
12.            & Techniques used for labeling           & Self-Assessment Mannequins (SAM)            \\
13.            & Labeling data scale                    & Continuous data values with a range of 0-9. \\
14.            & Labeling Classes                       & Arousal, Valence                            \\ \bottomrule
\end{tabular}%
}

\end{table}

\subsection{Electrode Selection}

The 10/20 System, also known as the International 10/20 System, is the standard used to position the electrodes on the scalp. We have used all the 32 channels for analysis in this study. 

\begin{figure}[!ht]
\centering
\includegraphics[scale = 0.25]{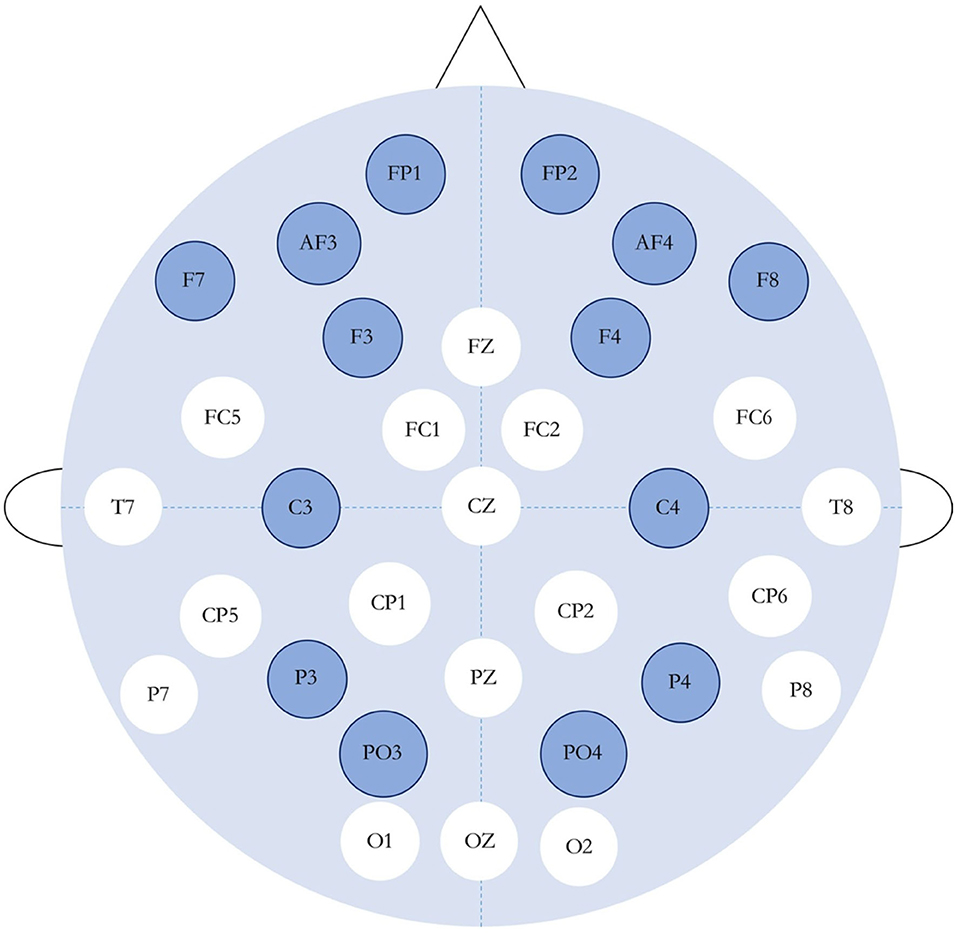}
\caption{10-20 Electrode Placement Technique} \label{fig1}
\end{figure}

\section {Methodologies and Experiments}

\subsection{Pre-Processing}

The DEAP EEG signals are already pre-processed so as to remove noise and artifacts. According to the data-set description webpage (DEAP, 2012), the original data was down-sampled to 128 Hz and the artifacts resulting from eye movements were removed with a blind source separation technique, as described in \cite{r24}. In order to keep the frequencies of interest, a band-pass frequency filter from 4.0-45.0 Hz was applied, and finally the data was averaged to the common reference.

\subsection{Exploration and Visualization}

\begin{figure}[!ht]

\centering
\includegraphics[scale = 0.6]{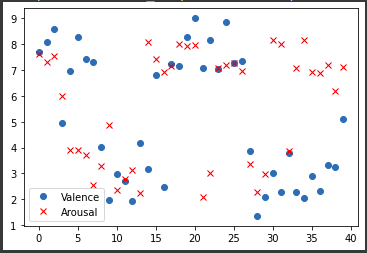}
\caption{Plot for first 40 rows (first participant)}
\end{figure}

The data set consists of 32 Matlab files for each participant. We loaded Matlab files using SciPy module and converted respective data and labels into arrays which have been reshaped into desired shapes. The participants rated the videos based on the Arousal-Valence, like/dislike, and dominance. In this work, we have considered only Arousal and Valence for labels. We have appended the data for 22 participants with respective labels for training purposes which results in 880 training samples (each participant with 40 trials). 
Labels:  (1280, 4)
Data:  (1280, 40, 8064)

While the cumulative mean of ratings of all the videos watched by a single participant, the trend line for valence and arousal gives a standard deviation difference of 0.57.

\begin{figure}[!ht]
\caption{Plot of valence and arousal rating between groups}
\centering
\includegraphics[scale = 0.6]{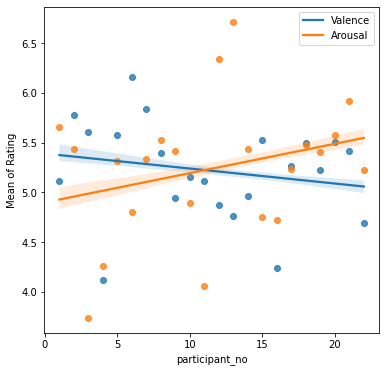}
\end{figure}

We have divided the combinations of valence and arousal into 4 groups (emotional states): High Arousal Positive Valence (Excited, Happy), Low Arousal Positive Valence (Calm, Relaxed), High Arousal Negative Valence (Angry, Nervous), and Low Arousal Negative Valence (Sad, Bored). The groups have been divided based on median values of Valence and Arousal ratings and the number of trials per each group has been calculated.

\subsubsection{Separating EEG \& Non-EEG data}

Data provided in Matlab files consists of psychological signals from 32 main channels and other peripheral channels. We have separating EEG Signals obtained from 32 channels named:
\newline
\newline
["Fp1", "AF3", "F3", "F7", "FC5", "FC1", "C3", "T7", "CP5", "CP1", "P3", "P7", "PO3", "O1", "Oz", "Pz", "Fp2", "AF4", "Fz", "F4", "F8", "FC6", "FC2", "Cz", "C4", "T8", "CP6", "CP2", "P4", "P8", "PO4", "O2"]

\subsection{Feature Extraction}

It is difficult to extract useful information from EEG signals just by observing them in the time domain. Therefore, there are many advanced signal processing techniques that can be used to extract relevant features from those signals. The choice of a particular technique is usually tied to the application under study and specific requirements. Feature extraction aims at describing relevant information about the brain activity by an ideally small number of relevant values. All extracted features are usually arranged into a vector, known as a feature vector, which is used later for brain activity classification.
There are three main sources of information that can be extracted from EEG readings:
spatial information (for multichannel EEG), spectral information (power in frequency
bands), and temporal information (time windows-based analysis).
\paragraph{Fourier analysis} is a popular signal processing approach to go from time domain signals to frequency domain signals or vice versa. This analysis can be applied to both continuous and discrete time signals. It relays on the principle that every signal can be represented or approximated by sums of trigonometric functions. FFT is an algorithm that computes the Discrete Fourier Transform (DFT) of a sequence, or its inverse. It produces the exact same result as evaluating the DFT definition directly with the difference that is much faster.

For the feature extraction from EEG data, we have used Welch's method (a part of Fourier analysis) to extract theta, alpha, beta, and gamma spectral power for each electrode. The frequency bands used: theta (4 - 8 Hz), alpha (8 - 12 Hz), beta (12 - 30 Hz), and gamma (30 - 64 Hz).

\begin{figure}[!ht]
\caption{Welch's periodogram for four frequency bands}
\centering
\includegraphics[scale = 0.6]{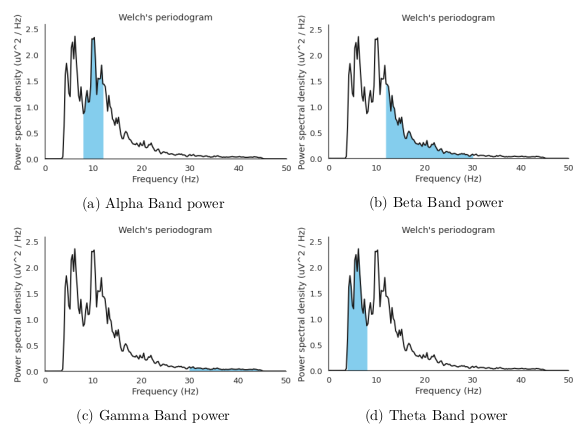}
\end{figure}

\subsection{Classifiers}
The PSD band power values were used in different combinations of channels and frequency bands for classification purposes in arousal and valence dimensions. To access the association between these states and EEG, the machine learning classifiers used were Support Vector Machine(SVM), K Nearest Neighbors (KNN), and a Multi Layer Perceptron. The objective of SVM is to find a hyperplane in N-dimensional space(N — the number of features) that distinctly classifies the data points. Whereas the KNN uses proximity to make classifications or predictions about the grouping of an individual data point. In MLP, the neurons use non-linear activation functions that imitate the behavior of the neurons in the human brain. It has a linear activation function in all its neurons and uses back-propagation for its training.

The classification scheme also includes a Tabular learning approach i.e. Google's TabNet. \cite{r25} TabNet is a deep neural network specifically designed to learn from tabular data, where it selects relevant features are selected at each decision step in the encoder using sparse learned masks. It is to ensure the selection of a small subset of features which makes model learning more efficient as model capacity is fully utilized at each decision step. Instead of using a harsh threshold on a feature, TabNet uses these learnable masks to make a soft decision. In this approach, the 10-fold cross-validation method was for classification, taking the average of 10 tests as the final classification results.

\section{Results and Discussion}

The cross-validation results were obtained after training and testing on the EEG data after extracting PSD in different frequency bands namely alpha, beta, theta, and gamma against two levels of Valence and two levels of Arousal. Test accuracies were compared against different channels, described as:
left ["Fp1", "AF3", "F7", "FC5", "T7"], right ["Fp2", "AF4", "F8", "FC6", "T8"], frontal ["F3", "FC1", "Fz", "F4", "FC2"], parietal ["P3", "P7", "Pz", "P4", "P8"], occipital ["O1", "Oz", "O2", "PO3", "PO4"], central ["CP5", "CP1", "Cz", "C4", "C3", "CP6", "CP2"]. Table 2 and Table 3 show the test accuracies of KNN, SVM and MLP against Arousal (High and Low Arousal) and Valence ( Positive and Negative Valence). Below are the highest scores obtained across the classifiers:

For arousal, 
    \begin{itemize}
    \item
    Top 3 regions with the highest scores are central (64.23), parietal (64.23) and frontal (61.79) and Top 2 bands with the highest scores are theta (64.23) and beta (64.23) for KNN classifier. 
    \item
    Top 3 regions with the highest scores are left (61.79), parietal (60.16) and central (54.47) and Top 2 bands with the highest scores are gamma (64.23) and alpha (64.23) for SVM classifier. 
    \item
    Top 3 regions with the highest scores are left (60.16), parietal (60.16) and central (57.72) and Top 2 bands with the highest scores are gamma (60.16) and theta (59.35) for MLP classifier. 
    \end{itemize}

For Valence,
    \begin{itemize}
    \item
    Top 3 regions with the highest scores are left (69.11), right (68.29) and central (67.48) and Top 2 bands with the highest scores are beta (69.11) and gamma (68.29) for KNN classifier. 
    \item
    Top 3 regions with highest scores are central (60.16), occipital (60.16) and frontal (58.45) and Top 2 bands with the highest scores are gamma (60.16) and theta (46.23) for SVM classifier. 
    \item
    Top 3 regions with the highest scores are frontal (56.91), central (55.28) and left (52.85) and Top 2 bands with the highest scores are gamma (56.91) and theta (54.47) for MLP classifier. 
    \end{itemize}

TabNet on the other hand, when trained across all 32 channels and features from all frequency bands ( alpha, beta, gamma, delta) against the labels High Arousal High Valence, High Arousal Low Valence,  Low Arousal High Valence, Low Arousal Low Valence, produced the test accuracy of 98.86 percent. TabNet is a complex model composed of a feature transformer, attentive transformer, and feature masking,  that soft feature selection with controllable sparsity in end-to-end learning. The reason for the high performance of TabNet is because it focuses on the most important features that have been considered by the Attentive Transformer. The Attentive Transformer performs feature selection to select which model features to reason from at each step in the model and a Feature Transformer processes features into more useful representation and to learn complex data patterns which improve interpretability and help it learn more accurate models.

\begin{table}[ht]
\centering
\caption{Accuracies of Different Classifiers for Arousal}
\resizebox{\textwidth}{!}{
\begin{tabular}{@{}llcccccclcccccclcccccc@{}}
\toprule
               &  & \multicolumn{6}{c}{\textbf{KNN}}                                                                              &  & \multicolumn{6}{c}{\textbf{SVM}}                                                                              &  & \multicolumn{6}{c}{\textbf{MLP}}                                                                              \\ \cmidrule(r){1-1} \cmidrule(lr){3-8} \cmidrule(lr){10-15} \cmidrule(l){17-22} 
\textit{Bands} &  & \textbf{left} & \textbf{frontal} & \textbf{right} & \textbf{central} & \textbf{parietal} & \textbf{occipital} &  & \textbf{left} & \textbf{frontal} & \textbf{right} & \textbf{central} & \textbf{parietal} & \textbf{occipital} &  & \textbf{left} & \textbf{frontal} & \textbf{right} & \textbf{central} & \textbf{parietal} & \textbf{occipital} \\ \midrule
\textbf{Theta} &  & 55.28         & 60.16            & 56.1           & 64.23            & 58.54             & 55.28              &  & 51.22         & 47.97            & 50.41          & 52.85            & 56.1              & 51.22              &  & 52.85         & 54.47            & 51.22          & 56.1             & 59.35             & 50.41              \\
\textbf{Alpha} &  & 51.22         & 56.1             & 50.41          & 59.35            & 57.72             & 51.22              &  & 53.66         & 50.41            & 50.41          & 52.03            & 60.16             & 51.22              &  & 51.22         & 56.1             & 51.22          & 57.72            & 57.72             & 51.22              \\
\textbf{Beta}  &  & 52.03         & 55.28            & 52.03          & 55.28            & 64.23             & 55.28              &  & 53.66         & 53.66            & 52.03          & 54.47            & 59.35             & 52.85              &  & 55.28         & 50.41            & 47.15          & 52.85            & 58.54             & 47.15              \\
\textbf{Gamma} &  & 57.72         & 61.79            & 58.54          & 56.91            & 61.79             & 58.54              &  & 61.79         & 51.22            & 52.03          & 53.66            & 59.35             & 52.03              &  & 60.16         & 47.97            & 45.53          & 56.1             & 60.16             & 43.09              \\ \bottomrule
\end{tabular}}

\end{table}
\vspace{-15mm}
\begin{table}[ht]
\centering
\caption{Accuracies of Different Classifiers for Valence}
\resizebox{\textwidth}{!}{
\begin{tabular}{@{}llcccccclcccccclcccccc@{}}
\toprule
               &  &               &                  &                & \textbf{KNN}     &                   &                    &           &               &                  &                & \textbf{SVM}     &                   &                    &           &               &                  &                & \textbf{MLP}     &                   &                    \\ \cmidrule(r){1-1} \cmidrule(lr){3-8} \cmidrule(lr){10-15} \cmidrule(l){17-22} 
\textit{Bands} &  & \textbf{left} & \textbf{frontal} & \textbf{right} & \textbf{central} & \textbf{parietal} & \textbf{occipital} & \textbf{} & \textbf{left} & \textbf{frontal} & \textbf{right} & \textbf{central} & \textbf{parietal} & \textbf{occipital} & \textbf{} & \textbf{left} & \textbf{frontal} & \textbf{right} & \textbf{central} & \textbf{parietal} & \textbf{occipital} \\
\textbf{Theta} &  & 56.91         & 63.41            & 57.72          & 67.48            & 56.91             & 55.28              &           & 46.34         & 43.09            & 41.46          & 39.84            & 46.34             & 43.9               &           & 42.28         & 54.47            & 46.34          & 45.53            & 43.09             & 42.28              \\
\textbf{Alpha} &  & 58.54         & 52.85            & 60.16          & 65.85            & 62.6              & 55.28              &           & 46.34         & 40.65            & 42.28          & 39.84            & 45.53             & 43.9               &           & 47.97         & 51.22            & 49.59          & 46.34            & 41.46             & 43.9               \\
\textbf{Beta}  &  & 69.11         & 64.23            & 65.85          & 59.35            & 64.23             & 63.41              &           & 46.34         & 43.09            & 45.53          & 44.72            & 42.28             & 45.53              &           & 47.97         & 43.09            & 48.78          & 52.85            & 52.85             & 47.97              \\
\textbf{Gamma} &  & 59.35         & 61.79            & 68.29          & 61.79            & 60.98             & 62.6               &           & 52.03         & 58.54            & 46.34          & 60.16            & 47.15             & 60.16              &           & 52.85         & 56.91            & 46.34          & 55.28            & 52.03             & 52.03              \\ \bottomrule
\end{tabular}}

\end{table}

\section{Conclusion}

This study focuses on analysis of DEAP Dataset and classifying the antithetical emotions with the best fit classifier. Using Welch’s Method of Fast Fourier Transform, Power Spectrum Density in different frequency bands has been used as a feature for training the classifiers. SVM, KNN, and MLP have been trained on all the 32 channels divided into six EEG regions and four frequency bands. KNN has given marginally better results than SVM and MLP on both Valence and Arousal datasets. As the features of all channels across all the frequency bands when presented into a single tabular data against four lables, namely High Arousal High Valence, Low Arousal High Valence, High Arousal High Valence, Low Arousal Low Valence, TabNet gave quite impressive results. The key benefit TabNet has is its explainability, as it uses instance-wise feature selection using masks in its encoder. This allows the model’s learning capacity to be focused on important features. It can capitalize on the forthcoming visualization of masks (as it provides explainability) which can be experimented in the future and which features are being used at a prediction level can be explored as a part of future work.  

\bibliographystyle{splncs04_unsort}
\bibliography{refrences.bib}

\end{document}